%% file: sample-sigconf.tex
\documentclass[sigconf]{acmart}

\AtBeginDocument{%
  \providecommand\BibTeX{{%
    \normalfont B\kern-0.5em{\scshape i\kern-0.25em b}\kern-0.8em\TeX}}}

\copyrightyear{2024}
\acmYear{2024}
\setcopyright{rightsretained}
\acmConference[GECCO '24 Companion]{Genetic and Evolutionary Computation
Conference}{July 14--18, 2024}{Melbourne, VIC, Australia}
\acmBooktitle{Genetic and Evolutionary Computation Conference (GECCO '24
Companion), July 14--18, 2024, Melbourne, VIC, Australia}
\acmDOI{ 10.1145/3638530.3654356}
\acmISBN{979-8-4007-0495-6/24/07}

\copyrightyear{2024}
\acmYear{2024}
\setcopyright{rightsretained}
\acmConference[GECCO '24]{Genetic and Evolutionary Computation Conference}{July 14--18, 2024}{Melbourne, VIC, Australia}
\acmBooktitle{Genetic and Evolutionary Computation Conference (GECCO '24), July 14--18, 2024, Melbourne, VIC, Australia}
\acmDOI{10.1145/3638529.3654113}
\acmISBN{979-8-4007-0494-9/24/07}
\usepackage{printlen}
\begin{document}

%%
%% The "title" command has an optional parameter,
%% allowing the author to define a "short title" to be used in page headers.
\title{Growing Artificial Neural Networks for Control:\\ the Role of Neuronal Diversity}

%%
%% The "author" command and its associated commands are used to define
%% the authors and their affiliations.
%% Of note is the shared affiliation of the first two authors, and the
%% "authornote" and "authornotemark" commands
%% used to denote shared contribution to the research.

\author{Eleni Nisioti$^{1}$\footnotetext[1]{Co-first author}, Erwan Plantec$^{1}$, Milton Montero, Joachim Winther Pedersen, Sebastian Risi}
\email{ {enis, erpl, mile, jwin, sebr}@itu.dk, }
\orcid{000-0001-7170-7108}
\affiliation{%
  \institution{IT University Copenhagen}
  \streetaddress{}
  \city{Copenhagen}
  \state{}
  \country{Denmark}
  %\postcode{43017-6221}
}

%%
%% By default, the full list of authors will be used in the page
%% headers. Often, this list is too long, and will overlap
%% other information printed in the page headers. This command allows
%% the author to define a more concise list
%% of authors' names for this purpose.
\renewcommand{\shortauthors}{.}

%%
%% The abstract is a short summary of the work to be presented in the
%% article.
\begin{abstract}

In biological evolution complex neural structures grow from a handful of cellular ingredients.
As genomes in nature are bounded in size, this complexity is achieved by a growth process where cells communicate locally to decide whether to differentiate, proliferate and connect with other cells. This self-organisation is hypothesized to play an important part in the generalisation, and robustness of biological neural networks. Artificial neural networks (ANNs), on the other hand, are traditionally optimized in the space of weights. Thus, the benefits and challenges of growing artificial neural networks remain understudied. Building on the previously introduced Neural Developmental Programs (NDP), in this work we present an algorithm for growing ANNs that solve reinforcement learning tasks.
We identify a key challenge: ensuring phenotypic complexity requires maintaining neuronal diversity, but this diversity comes at the cost of optimization stability. To address this, we introduce two mechanisms: (a) equipping neurons with an intrinsic state inherited upon neurogenesis; (b) lateral inhibition, a mechanism inspired by biological growth, which controlls the pace of growth, helping diversity persist.
We show that both mechanisms contribute to neuronal diversity and that, equipped with them, NDPs achieve comparable results to existing direct and developmental encodings in complex locomotion tasks.

\end{abstract}


\begin{teaserfigure}
    \includegraphics[width=\textwidth]{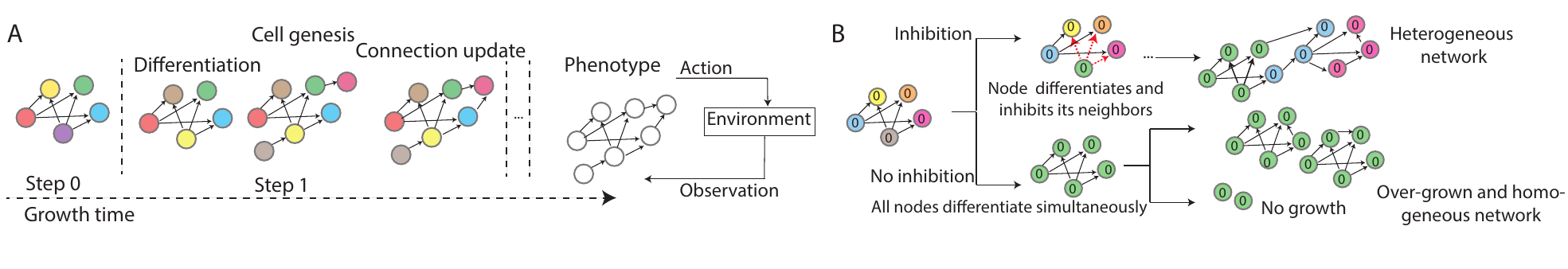}
\caption{Neural Developmental Programs with Lateral Inhibition. \normalfont (A) Growth starts with a minimal number of cells with unique intrinsic hidden states (indicated with numbers) and random extrinsic hidden states (indicated with colors). At each step, each cell independently decides whether to differentiate (update its extrinsic hidden state based on its environment), generate a new cell that inherits its intrinsic state and update a connection. Afterward, the grown network is mapped to a recurrent neural network used to control an agent in an environment. (B) Lateral inhibition: each time a cell differentiates, grows a new cell or updates a connection,  it inhibits its neighbors for a fixed number of steps. Inhibition helps maintain diversity in the intrinsic hidden states.}
\label{fig:method}
\end{teaserfigure}

%% A "teaser" image appears between the author and affiliation
%% information and the body of the document, and typically spans the
%% page.
%\begin{teaserfigure}
%  \includegraphics[width=\textwidth]{sampleteaser}
%  \caption{Seattle Mariners at Spring Training, 2010.}
%  \Description{Enjoying the baseball game from the third-base
%  seats. Ichiro Suzuki preparing to bat.}
%  \label{fig:teaser}
%\end{teaserfigure}

%%
%% This command processes the author and affiliation and title
%% information and builds the first part of the formatted document.
\maketitle

\input{sections/introduction}

\input{sections/methods}

\input{sections/results.tex}

\input{sections/discussion.tex}

\section*{Acknowledgments}
This project was supported by a European Research Council (ERC) grant (GA no.
101045094, project ”GROW-AI”).

\bibliographystyle{ACM-Reference-Format}
\bibliography{sample-base}
\appendix

\end{document}

%% file: sections/introduction.tex
Biological neural networks, a family that exemplifies some of the most complex systems in life, are the products of a growth process that starts with a handful of cells. 
To support complex behavior, these cells proliferate, differentiate and connect with others, forming networks consisting of thousands of different cell types~\citep{sepp_cellular_2024,sanes_development_2007}.
They do so with limited information.
Genomes are not large enough to directly encode brain states ~\citep{zador_critique_2019,koulakov_encoding_2022} and cells can only perceive information locally.
Artificial neural networks (ANNs), on the other hand, follow an engineering rather than a biological paradigm: the architecture and weights are optimized directly and transferred intactly to the next generation~\citep{stanley_taxonomy_2003,stanley_evolving_2002,salimans_evolution_2017}.

Inspired by the genotype-to-phenotype mapping found in nature, on the other hand, artificial developmental encodings optimize ANNs in a search space of lower dimensionality and define a mapping from this space to the ANN realisation~\citep{schmidgall_brain-inspired_2023,ha_hypernetworks_2016,stanley_hypercube-based_nodate,najarro_towards_2023}. 
This process of retrieving the ANN from can be seen as a model of biological development.
Some developmental encodings abstract away the temporally-extended nature of development and formulate it as a single-step, non-linear mapping~\citep{stanley_hypercube-based_nodate,ha_hypernetworks_2016}, while others explicitly model the self-organised nature of growth using primarily Cellular Automata~\citep{mordvintsev2020growing}.
Reported benefits of such developmental encodings are: a) improved generalisation due to the regularising efffect of a genomic bottleneck\citep{zador_critique_2019,koulakov_encoding_2022,stanley_hypercube-based_nodate} b) structural patterns such as modularity and regularity that naturally emerge from self-organised processes and can improve, among others, the energy efficiency of phenotypes  \citep{stanley_hypercube-based_nodate} c) robustness to perturbations that may occur during growth when the latter is self-organised~\citep{mordvintsev2020growing}.

In this work we study Neural Developmental Programs (NDPs), a previously proposed algorithm for growing ANNs~\citep{najarro_towards_2023} that control the behavior of an agent in RL tasks.
NDPs determine both the architecture and weights of the ANN through a temporally-extended, self-organised growth process.
Inspired by biological growth, they equip cells with the ability to differentiate, grow new neurons and update their synaptic weights.
All cells share a model that determines how these decisions are made based on locally available information.
Compared to morphogenetic algorithms that leverage a target at each step of growth~\citep{mordvintsev2020growing}, the RL setting introduces additional challenges to the stability and success of development: the growth process does not have access to a target control policy but only receives some feedback regarding its performance.
Compared to one-step indirect encodings for control~\citep{stanley_hypercube-based_nodate,ha_hypernetworks_2016},
NDPs employ a more complex genotype-to-phenotype mapping that increases training difficulty, but introduces the potential of leveraging the benefits of self-organisation for control.

We empirically identify a challenge with training NDPs: if all neurons differentiate into the same type, then growth-related decisions will be identical.
ANNs grown by such a naive process have a homogeneous structure and cannot exhibit complex behavior.
Based on this observation we propose two modifications.
First, employing intrinsic states that are not modified during growth ensures that diversity cannot disappear in the population.
As networks are initialized with a small number of cells that each have a unique intrinsic state, some diversity is introduced at the beginning of growth.
As the networks grow, these unique states are copied around so that not every cell has a unique intrinsic state.
Thus, similarly to biological networks, lineages of cells are formed during growth.
As we show empirically, the introduction of intrinsic states can make the difference between complete failure and success in locomotion tasks.
Second, we introduce \textit{lateral inhibition}, a mechanism hypothesized to play an important role in the maintenance of diversity in biological growth~\citep{sanes_development_2007,ables_notch_2011,fares_cooperative_2009}.
Under this mechanism, when a cell decides to perform an action it inhibits other cells in its neighborhood from performing a similar action for a limited number of steps.
Our simulations indicate that inhibition helps maintain neuronal diversity both during growth and during evolution, even in the absence of intrinsic states.

%% file: sections/methods.tex
\section{Methods}\label{sec:model}

\subsection{The Neural Developmental Program}

We model growth as a stochastic graph generation process whose final output is the ANN used to control the actions of an RL agent.
At each growth step the state of the agent is represented by a directed graph ($\mathcal{G}$) where vertices ($\mathcal{V}$) represent cells and edges ($\mathcal{E}$) connections between them.
Cells and edges are characterized by real-valued vectors, termed their hidden states ($h$ and $e$ for cells and edges respectively).
The hidden state of a cell consists of two parts: $h_{\text{ext}}$ is the extrinsic hidden state, influenced by its environment, and $h_{\text{int}}$ is the intrinsic hidden state that only depends on its lineage (to be defined shortly).

Growth starts with a graph that contains a number of cells equal to the size of the observation plus action space of the environment, as this is the minimum number of neurons that the policy network needs to have. 
Each cell in this initial graph has a unique intrinsic state. At each step a cell can perform three actions:
\begin{itemize}
    \item \textit{differentiate}, i.e., predicts its next hidden state using a model that takes into account the local environment of the cell.
    We refer to this model as the DiffModel and implement it as a Graph Attention Network (GAT) \citep{velickovic_graph_2018} that takes as input the hidden states of all nodes in the graph and the current connectivity.
    In general different implementation of the DiffModel are possible, such as graph convolution employed in the original introduction of NDPs~\citep{najarro_towards_2023}.
    \item \textit{grow}, i.e., generate a new cell. To decide whether to grow a cell employs another model, the GenModel $G(h) \rightarrow \{0,1\}$ that we implement as a feedforward network  shared across neurons. The newly generated cell forms a connection with its parent and inherits its intrinsic state. Thus, lineages of cells are formed during growth.
\item \textit{update the weight of a connection}. The new value of a  weight depends on the hidden state of the two cells connected to it and is predicted by a third model, the EdgeModel $E(h_1,h_2) \rightarrow \mathcal{R}$, that is also a feedforward network.
\end{itemize}
Thus, the growth process is orchestrated by the DiffModel, GenModel and EdgeModel, which can be seen as the developmental program that runs in each cell of an organism.
At the end of the growth process, which lasts for a fixed number of steps, the ANN has acquired its final connectivity.
To derive the behavioral policy we interpret the cells of the final graph as neurons, the weights of the edges as the weights of the ANN, and equip neurons with fixed activation functions to form a recurrent ANN.
We present a schematic of our model in Figure \ref{fig:method}. 
To optimize our model for a given task we train the weights of the three models in the NDP using an evolutionary strategy.
Thus, the number of optimization parameters does not depend on the size of the control policy.

%Similarly to previous works for growing ANNs, we refer to them as a Neural Developmental Program (NDP) ($(D, G, E)$)\citep{najarro_towards_2023}.

%To differentiate between the graphs used during growth and the final graph we refer to the former as \textit{substrate graphs} and to the latter as the \textit{target graph}.

%As we will discuss in our experiments, NDPs of small size are adequate for growing complex locomotion policies.

\subsection{Maintaining neuronal diversity}

Biological growth is characterized by a \textit{diversity paradox}: a handful of identical cells grow to become a network consisting of thousands of different neuron types~\citep{sanes_development_2007}.
If neurons change only based on local information and if they all start off with the same environment, then where does this diversity come from? 
As we show in our experimental analysis, the same question becomes relevant when growing ANNs.
Although the DiffModel could theoretically learn to differentiate neurons so that their hidden states remain diverse, this seems hard to achieve in practice.
Rather, the hidden states converge to identical values early in the growth process, leading to degenerate solutions where all weights explode to large values or effectively disappear.

To avoid this instability we turn towards a mechanism hypothesized to play an important role in maintaining diversity in biological growth: \textit{lateral inhibition}~\citep{sanes_development_2007}.
Under this mechanism, a cell that undergoes a certain change, such as differentiation, neurogenesis or synaptogenesis, emits a signal to its neighborhood that prohibits other cells from undertaking a similar action for a fixed number of steps ~\citep{sanes_development_2007,ables_notch_2011,fares_cooperative_2009}.
This ensures that cells do not make decisions simultaneously.
We visualize the concept of of lateral inhibition in Figure~\ref{fig:method}. As we show in our experiments, both lateral inhibition and the use of intrinsic states contribute to the maintenance of diversity. 

%In our model, under a purposefully simplistic interpretation of this mechanism, a cell that undergoes a certain change, such as differentiation, neurogenesis or synaptogenesis, emits a signal to its neighborhood that prohibits other cells from undertaking a similar action for a fixed number of steps ~\citep{sanes_development_2007,ables_notch_2011,fares_cooperative_2009}.

%% file: sections/results.tex
\section{Results}\label{sec:results}

\paragraph{Experimental setup} We test the ability of NDPs to learn how to solve complex behavioral tasks. 
In particular, we experiment with the Reacher, Inverted Double Pendulum, Halfcheetah and Ant tasks using the Mujoco Brax library~\citep{brax2021github}.
The three latter tasks evaluate for locomotion while Reacher in addition evaluates the ability to locate different targets.
%In all experiments with the NDP, substrate networks start with the minimal number of cells required to grow a target network that has all the input and output neurons for the corresponding environment.
The intrinsic hidden states are one-hot encoded vectors of length equal to the number of initial cells.
The extrinsic hidden states are real-valued vectors of length 8.
The EdgeModel is a feedforward network with two hidden layers of 16 neurons each with a ReLU activation function for hidden neurons and a linear activation function for output neurons.
The GenModel is a feedforward network with one hidden layers of 32 neurons with a ReLU activation function for hidden neurons and a tanh activation function for the output neuron.
The target network used to control the policy employs a ReLU activation function for hidden neurons and a linear one for the output layers.
Growth lasts for 15 steps and  inhibition lasts for 2 steps.
All methods are trained with DES \citep{lange_discovering_2023}, which we empirically found to perform better than other evolutionary strategies, using the Evosax library \citep{evosax2022github}.
Results are averaged across 3 training trials and 10 evaluation trials.
We provide our code for running end-to-end training on a GPU and visualizations of the learned trajectories in \href{https://github.com/eleninisioti/GrowNeuralNets}{an online repo}.

%We empirically show that the NDP is prone to learning a trivial growth policy that does not grow any hidden neurons. 
%This happens when we do not employ either intrinsic hidden states or lateral inhibition.
%We refer to this version of the NDP as NDP-vanilla and show that the reason behind its degeneration is the loss of neuronal diversity at the early stages of growth.
Our main result is that the NDP with intrinsic hidden states discovers optimal policies for all tasks.
We compare its performance to two baselines: a) a direct encoding where DES directly optimizes the weights of an RNN with 100 neurons (which corresponds to the maximum size of networks grown by the NDP) b) an indirect encoding that employs an EdgeModel to predict weights based on the intrinsic state of each neuron in a single step.
This method can be seen as a Hypernetwork \citep{ha_hypernetworks_2016} with a pre-determined (rather than learned) encoding of target neurons.
Thus, this ablation differs from our NDP in that there are no extrinsic hidden states, the network architecture is fixed and all neurons have a unique intrinsic state.
Our objective is to show that the NDP performs comparably to the indirect encoding despite the training difficulties we have discussed.
The direct encoding is provided as a baseline that is known to perform well in the tasks we examine and, therefore, serves as an expected upper threshold for fitnesses.
As these tasks do not pose any particular need for generalisation or robustness we do not expect to see indirect encodings outcompeting direct ones.

\begin{figure*}
    \includegraphics[width=\textwidth]{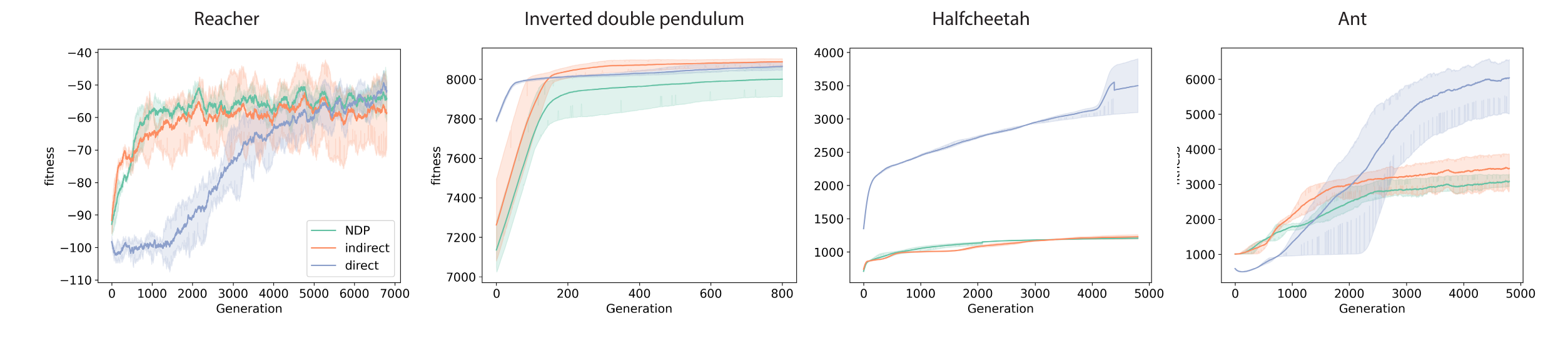}
    \caption{Training curves for different control tasks and methods. NDPs perform comparably to the other indirect encoding.}
 \label{fig:overall}
\end{figure*}

%\paragraph{NDP performs on par with direct encodings}
Figure \ref{fig:overall} presents the evaluation performance during training, where every 10 generations we perform 2 evaluation trials of the currently best individual.
We observe that the NDP performs comparably to the indirect encoding , as ANOVA indicates that differences are statistically insignificant for all tasks.
In contrast, evaluating the NDP without intrinsic states led to trivial policies that did not show any improvement in any of the tasks.
The direct encoding converges more slowly than the other methods for some tasks (Reacher and Ant), arguably due to the larger size of the search space.
In the Ant and Halfcheetah, the direct encoding achieves high speed much quicker than the indirect encoding and NDP, which progress slowly but eventually solve the task (in an additional experiment we saw that the NDP requires about 10000 generations to achieve a fitness of 6000 in the Ant).  
A possible explanation for this is that Halfcheetah and Ant have a larger action space compared to the other tasks and may, therefore, prove more challenging for indirect encodings or require retuning of the EdgeModel.

%Finally, we show that introducing lateral inhibition into the NDP helps maintain neuronal diversity and grow more heterogeneous target networks compared to solely employing intrinsic hidden states.

\begin{figure}
    \includegraphics[width=\columnwidth]{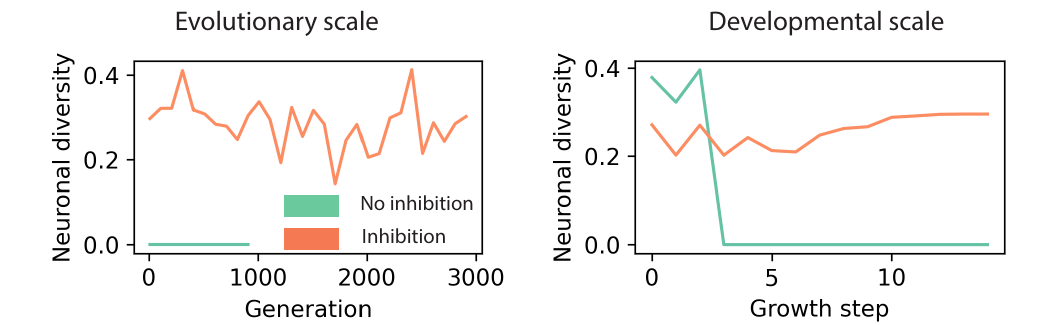}
    \caption{Effect of inhibition on neuronal diversity in the Reacher  during evolution (left) and during growth (right). }
 \label{fig:neural_div}
\end{figure}

%\paragraph{Inhibition helps maintain neuronal diversity}
%We now move towards evaluating the effect of inhibition.
In Figure \ref{fig:neural_div} we monitor neuronal diversity both during evolution (where we only look at the final growth step) and during growth (after convergence at the evolutionary scale). Neural diversity is defined as the average distance between a neurons' hidden states and the ones of its $k=10$ nearest neighbors.
We compare two conditions: NDP with lateral inhibition and NDP without lateral inhibition.
In both cases, we remove intrinsic states as they may affect neuronal diversity.
We observe that inhibition helps maintain neuronal diversity at both the evolutionary scale and developmental scale while it completely disappears within the first growth steps in the absence of inhibition.

%% file: sections/discussion.tex
\section{Discussion}
We studied Neural Developmental Programs, an algorithm inspired by biological growth and models the growth of an ANN as a stochastic graph generation process where decisions are made locally by interacting neurons.
We empirically showed that such algorithms are prone to training instabilities, which may explain why they have not been successfully applied to complex RL tasks in the past.
We identified neuronal diversity, the presence of neurons with different hidden states during growth, as an important step towards avoiding the degeneration of control policies.
%Without neuronal diversity, our algorithm grows ANNs with a small variance in their weights and often avoids growing any hidden neurons.
To ensure neuronal diversity we introduced two key ingredients: a) neurons start out with unique intrinsic hidden states that are inherited upon neurogenesis b) lateral inhibition is employed when neurons differentiate and generate new neurons.
Our additions to the NDP enabled it to go beyond the simple navigation and control tasks investigated in~\cite{najarro_towards_2023}  to complex 3D locomotion and manipulation tasks, demonstrating the important role of neural diversity in neural growth.

%We empirically show that both mechanisms contribute to increased neuronal diversity.

%Our work adds to the family of developmental encodings an algorithm that leverages self-organization that can be trained using an evolutionary strategy rather than requiring the computation of gradients.

Our work has been concerned with how to grow ANNs.
Another, equally urgent, question is why grow them.
Growth can be seen, especially in an engineering-focused deep-learning era, as a limitation rather than a beneficial feature of biological organisms.
Yet biologists, complexity- and computer science researchers have voiced a number of hypotheses on how growth can benefit an evolving system~\citep{Kauffman1993-KAUTOO-2,sanes_development_2007,kowaliw_growing_2014}.
Arguably the benefits of growth become more obvious once one introduces environmental feedback.
Then, similarly to developmental algorithms like Hebbian~\citep{shaw_donald_1986} and reinforcement learning, growth can render the organism adaptable to its environment.
Could growth be beneficial even in the absence of environmental input?
A relevant hypothesis is that evolution faces an upper threshold in the complexity of phenotypes it can create and that growth is a necessary ingredient for evolving systems of certain behavioral complexity~\citep{Kauffman1993-KAUTOO-2}.
Our current empirical analysis does not show any benefits of temporally-extended growth over the single-shot baseline.
We believe that such benefits for complexity and generalisation will appear if one scales up the tasks appropriately.
By stabilizing the training of NDPs to work in high-dimensional, continuous action spaces our work lays the ground for such studies. 
As the design space of NDPs is large, future work can investigate the effect that different implementations of differentiation, neurogenesis and synaptogenesis have on the training stability of NDPs and leverage additional mechanisms and biases from the study of growth in biology and computer science~\citep{pmlr-v188-maile22a,chandra_self-organized_2024,wolterhoff_synaptic_2024}.